\documentclass[11pt,a4paper]{article}
\usepackage{times,latexsym}
\usepackage{url}
\usepackage[T1]{fontenc}
\usepackage{booktabs}
\usepackage[table]{xcolor}
\usepackage{transparent}
\usepackage{url}
\usepackage{xcolor}

\usepackage[inline,shortlabels]{enumitem}

\usepackage{amsmath}
\usepackage{adjustbox}
\usepackage{inconsolata}
\usepackage{setspace} %

\usepackage[acceptedWithA]{tacl2021v1}

\usepackage{xspace,mfirstuc,tabulary}

\newif\iftaclinstructions
\taclinstructionsfalse %
\iftaclinstructions
\renewcommand{\confidential}{}
\renewcommand{\anonsubtext}{(No author info supplied here, for consistency with
TACL-submission anonymization requirements)}
\newcommand{\instr}
\fi

\iftaclpubformat %

\else

\fi

\newcommand{\gptfour}{GPT4-turbo}

\newcommand{\llama}{\texttt{Llama-3-8B-Instruct}}

\newcommand{\mixeval}{\texttt{MixEval-Hard}}

\newcommand{\ultrafeedback}{\texttt{UltraFeedback}}

\newcolumntype{H}{>{\setbox0=\hbox\bgroup}c<{\egroup}@{}}
\newcommand{\reviser}{\textit{Reviser}}
\newcommand{\judge}{\textit{Judge}}
\newcommand{\strongerpreferred}{Stronger Preferred}

\title{Anchored Preference Optimization and Contrastive Revisions: Addressing Underspecification in Alignment}

\author{Karel D'Oosterlinck$^{1,3}\thanks{~~Work done as a part of an internship at Contextual AI. Code at {\scriptsize \url{https://github.com/ContextualAI/CLAIR_and_APO}}}$\quad{}Winnie Xu$^3$\quad{}Chris Develder$^1$\quad{}Thomas Demeester$^1$ \\  \textbf{Amanpreet Singh$^3$\quad{}Christopher Potts$^2$\quad{}Douwe Kiela$^{2,3}$\quad{}Shikib Mehri$^3$} \\ \\ $^1$Ghent University -- imec\qquad$^2$Stanford University\qquad$^3$Contextual AI\\
\texttt{karel.doosterlinck@ugent.be},~\texttt{shikib@contextual.ai}
}

\date{}

\begin{document}
\maketitle

\begin{abstract}

Large Language Models (LLMs) are often aligned using contrastive alignment objectives and preference pair datasets.
The interaction between model, paired data, and objective makes alignment a complicated procedure, sometimes producing subpar results. 
We study this and find that (i)~preference data gives a better learning signal when the underlying responses are contrastive, and (ii)~alignment objectives lead to better performance when they specify more control over the model during training.
Based on these insights, we introduce Contrastive Learning from AI Revisions (CLAIR), a data-creation method which leads to more contrastive preference pairs, and Anchored Preference Optimization (APO), a controllable and more stable alignment objective.
We align \llama{} using various comparable datasets and alignment objectives and measure \mixeval{} scores, which correlate highly with human judgments. 
The CLAIR preferences lead to the strongest performance out of all datasets, and APO consistently outperforms less controllable objectives. Our best model, trained on 32K CLAIR preferences with APO, improves \llama{} by 7.65\%, closing the gap with \mbox{GPT4-turbo} by 45\%.
\end{abstract}

\section{Introduction} \label{sec:intro}

\begin{figure}
    \centering    \includegraphics[width=\columnwidth]{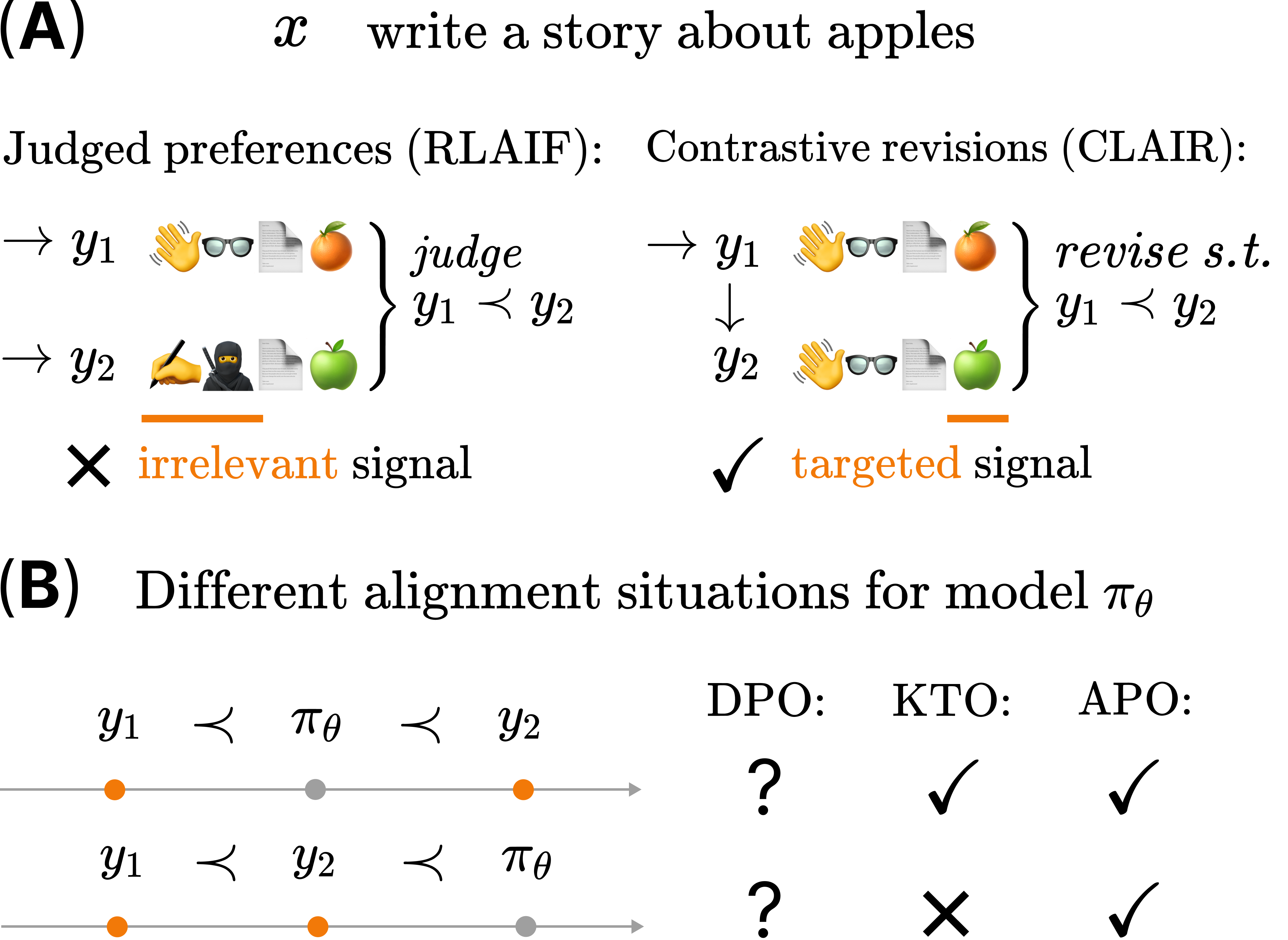}
    \caption{
    Alignment is underspecified with regard to preferences and training objective.
    \textbf{A:} Preference pairs can vary along irrelevant aspects, Contrastive Learning from AI Revisions (CLAIR) creates a targeted preference signal instead. 
    \textbf{B:} The quality of the model can impact alignment training, Anchored Preference Optimization (APO) explicitly accounts for this.
    }
    \label{fig:underspecified}
\end{figure}

Aligning language models with preferences
is a critical component in LLM development, significantly enhancing model capabilities, safety, and adherence to human values \citep{christiano2017deep, ouyang2022training, bai2022constitutional}.
These preferences can be expressed through \emph{preference pairs} (output $y_l \prec y_w$ for input $x$),
which offer a richer signal than individual outputs and enable more expressive learning objectives.
Recently, contrastive learning objectives have made alignment more accessible \citep{rafailov2024direct}.

Despite these advantages, alignment outcomes can be suboptimal
\citep{eisenstein2023helping, feng2024towards, park2024disentangling}.
In this paper, we reason through the nature of alignment, focusing on \begin{enumerate*}[(i)]
\item the preference signal expressed by the data, and
\item the training dynamics of contrastive objectives.
\end{enumerate*} 
We find that across both these axes, conventional alignment methods are underspecified.
To solve this, we argue that
\begin{enumerate*}[(i)]
    \item preference data should be minimally contrastive,
    and
    \item alignment objectives should account for distinct alignment situations
\end{enumerate*} (see Figure~\ref{fig:underspecified}).
This sheds light on suboptimal alignment outcomes. For example, we show in Section~\ref{sec:experiments} how a model aligned using high-quality outputs can actually degrade if the pairs differ in multiple uncontrolled aspects.

These insights lead to two new contributions. 
First, we introduce Contrastive Learning from AI Revisions (CLAIR), 
a method for creating preference pairs which \emph{minimally revises} one output to express a preference. 
The pairs created by CLAIR result in a more precise learning signal, as opposed to conventional methods which use a judge to \emph{select} a preferred response.
Second, we introduce Anchored Preference Optimization (APO), a family of contrastive objectives which explicitly account for distinct relationships between model and data during alignment.
The tailored training dynamics of APO results in more performant alignment compared to conventional objectives.

In order to study the role of both
\begin{enumerate*}[(i)]
        \item minimally contrastive preference data, and
        \item distinct alignment training dynamics,
\end{enumerate*}
we individually align a model across four comparable preference datasets using five alignment objectives.
One dataset is created through our CLAIR method. We compare this with two conventional judge-based datasets (Reinforcement Learning from AI Feedback; \citealt{bai2022constitutional}). Finally, we consider an 
ablated version of CLAIR created to directly assess the impact of contrastiveness.
We consider five distinct alignment objectives: DPO \citep{rafailov2024direct}, KTO \citep{ethayarajh2024kto}, continued Supervised Fine-Tuning on the preferred answer, and two variants of our proposed APO. 
We measure \mixeval{} accuracy \citep{ni2024mixeval} and length-controlled \texttt{AlpacaEval} scores \citep{dubois2024length} for each model, both benchmarks correlate highly with 
model rankings produced by humans \citep{chiangchatbot}.

We align \llama{} \citep{dubey2024llama} and use GPT4-turbo \citep{achiam2023gpt} for preference judgements\,/\,revisions. We find that our strongest model, aligned on 32K CLAIR preferences with APO, improves \llama{} performance by 7.65\% on \mixeval{}, closing the performance gap with GPT4-turbo by 45\%. Our analysis indicates that the contrastiveness of CLAIR preferences is the major driver of performance.
Across every alignment datasets considered, APO objectives achieve the best performance.
In our analysis, we outline how to select the best APO variant given a target model and preference dataset.
Finally, we deeply explore recent alignment efforts and discuss how they relate to CLAIR and APO.

\section{Underspecification in Alignment}\label{sec:factors}

The alignment procedure creates complex interactions between the target model, the preference dataset, and the alignment objective. The present section reflects on failure cases of all alignment efforts which start from preferences. The section discussed data and objective respectively.

Given a collection of prompts $X$, a preference dataset is a set of triples 
$(x, y_{w}, y_{l})$
, where $y_{w}$ and $y_{l}$ are, respectively, a winning (more preferred) and losing (less preferred) response to prompt~$x$. 
The preference signal in such a dataset is essentially expressed by the \emph{difference between} winning and losing outputs, illustrated in Figure~\ref{fig:underspecified}~A. 
However, paired outputs can differ in many aspects, some of which are spurious and thus irrelevant to the preference. 
These spurious differences will generally create a challenging credit assignment problem. 
Outputs which are \emph{minimally contrastive} differ along fewer axes, resulting in less spurious differences.
Thus, \textbf{if preference pairs produce a clearer minimal contrast, the alignment learning signal becomes more clear}.
Existing preference datasets vary  meaningfully in their contrastiveness. For example, in the Stanford Human Preferences dataset \citep{pmlr-v162-ethayarajh22a}, two outputs in a pair are simply responses to the same Reddit post, and thus they are not guaranteed to be especially comparable. 
An ideal preference dataset would consist of a very controlled difference between either example. This insight leads us to CLAIR (Section~\ref{sec:data}).

Preference triples only specify that one output is better than another.
This creates ambiguity, since it is not known if the more preferred answer was actually good. 
To see how this can impact alignment, suppose we have a dataset of triples
where $y_{w}$ tends to score 8/10 on some quality scale and $y_{l}$ tends to score 6/10. 
A target model that generally scores 9/10 may become worse if the likelihood of $y_{w}$ would increase during training, as illustrated in Figure~\ref{fig:underspecified}~B. %
Therefore, \textbf{alignment training needs to be aware of how desirable any individual answer is, regardless of its preference relationship}. 
To take a salient example, $\approx$80\% of winning outputs in \ultrafeedback{} \citep{cui2024ultrafeedback}
are generated by a less performant model than \llama{} (as measured by Chatbot Arena Elo; \citealt{chiangchatbot}). Naively aligning \llama{} on this dataset may thus worsen performance. 
Examples like this one
lead us to Anchored Preference Optimization (APO; Section~\ref{sec:objectives}).

\begin{figure*}[ht]
    \centering    \includegraphics[width=\textwidth]{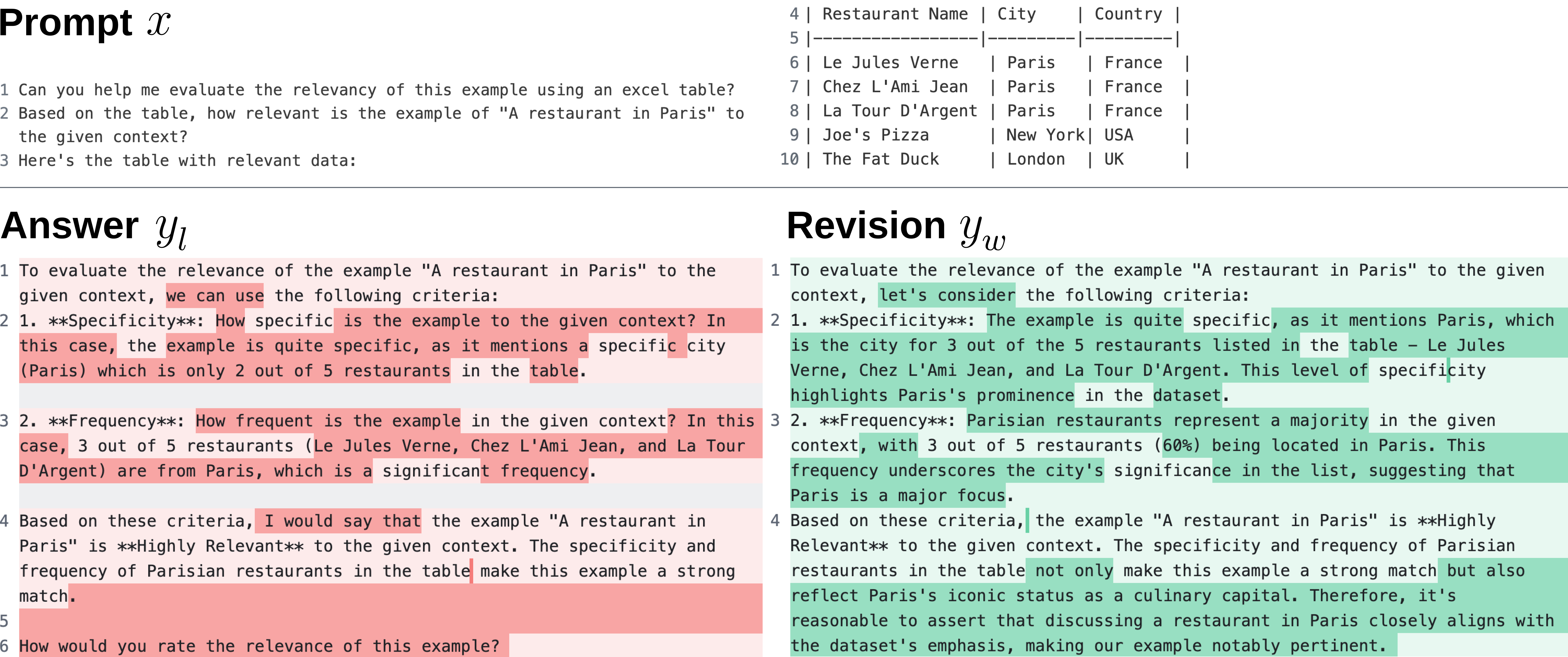}
    \caption{An \textbf{answer} produced by \llama{} for a \textbf{prompt}, and corresponding \gptfour{} \textbf{revision} of this answer. The differences between answer and revision are highlighted. The revision generally follows the same outline as the answer but improves it where possible. For example, the revision correctly alters the count of Parisian restaurants from 2 to 3 in the second line of the answer.}
    \label{fig:clair-example}
\end{figure*}

In summary, current alignment approaches are underspecified along two key axes:
\begin{enumerate*}[(i)]
    \item preferences may be weakly expressed due to non-contrastive data, and
    \item alignment objectives need to account for the model-data relation
\end{enumerate*}. 
In what follows, we set out to improve alignment across both axes.

\section{Contrastive Learning from Revisions} \label{sec:data}

We now introduce Contrastive Learning from AI Revisions (\textbf{CLAIR}), a general procedure for creating minimally contrasting preference pairs.%

Let $M$ be the target model we will align. 
Given a prompt $x$, we sample the losing output $y_{l}$ directly from the model. 
Then, we use a \reviser{} to minimally revise and improve $y_{l}$, resulting in the winning output $y_{w}$:
\begin{equation}
\begin{aligned}
y_{l} &= M(x) \\
y_{w} &= \reviser{}(x, y_{l}).
\end{aligned}\label{eq:clair}
\end{equation}
In this work, we use a stronger LLM to perform revisions, prompted to enhance the clarity, correctness, and engagement of the output (prompts and dataset details given in Appendix~\ref{app:prompts}).
Figure~\ref{fig:clair-example} shows an example 
triple created using this method. The losing output was generated by \llama{} and revised by \gptfour{}. The revision keeps most of the initial output intact, while improving details.
Recently, \citet{dubey2024llama} used human revisions in the development of the \texttt{llama-3.1} model family, though their process seems oriented towards enhancing quality differences rather than creating minimal contrasts.

CLAIR differs markedly from more familiar approaches to collecting preference data. For example,
in the \textbf{on-policy judge} paradigm (as used in Reinforcement Learning from AI Feedback; \citealt{bai2022constitutional}),
two generations are sampled from $M(x)$, and a \judge{} (often another LLM) decides which is the winner and which the loser:
\begin{equation}
\begin{aligned}
y_{1}, y_{2} &= M(x), M(x) \\ 
y_{w}, y_{l} &= \judge{}(x, y_{1}, y_{2}).
\end{aligned} \label{eq:rlaif}
\end{equation}
We use this approach as one of our baselines, with a prompt comparable to the revision prompt used by CLAIR.
Additionally, we consider an \textbf{off-policy judge} versions of \eqref{eq:rlaif}
    where the outputs are generated by models other than the target model:
\begin{equation}
\begin{aligned}
y_{1}, y_{2} &= M'(x), M''(x) \\ 
y_{w}, y_{l} &= \judge{}(x, y_{1}, y_{2}).
\end{aligned} \label{eq:offpol-rlaif}
\end{equation}

Both the on-policy and off-policy judge approaches provide useful comparison points for CLAIR.
In addition, we evaluate a baseline that helps us understand the role of contrastiveness in particular.  For CLAIR, the \reviser{} is generally a stronger model than the model we are aligning. This means that the winning examples $y_{w}$ are always generated by a stronger model. To decouple this factor from the contrastiveness induced by the revision process, we also evaluate a baseline that we call \textbf{\strongerpreferred}, where the stronger model provides the winning example for each pair without revision:
\begin{equation}
\begin{aligned}
    y_{l} &= M(x) \\ 
    y_{w} &= \textit{Stronger}(x) 
\label{eq:student-teacher}
\end{aligned}
\end{equation}

For the alignment experiments reported in Section~\ref{sec:experiments}, we created four preference datasets following \eqref{eq:clair}--\eqref{eq:student-teacher}.
Each dataset is created using the same 32K prompts uniformly sampled from \ultrafeedback{} \citep{cui2024ultrafeedback}, a widely used preference dataset with prompts spanning a broad range of domains. We take the target model $M$ to be \llama{}, one of the most competitive open source models available at the time of writing. 
For the off-policy judge dataset, we use already judged outputs available in \ultrafeedback{}. Approximately 80\% of these winning outputs are generated by a model weaker than \llama{} (as measured by Chatbot Arena Elo; \citealt{chiangchatbot}). Thus, this off-policy judge dataset generally contains lower quality outputs compared to the model.

Part of the goal of Section~\ref{sec:experiments} is to study the behavior of each of these datasets in the context of alignment efforts. However, one of the high-level goals of CLAIR is to generate examples that are minimally contrastive. We can assess this directly using some simple heuristics: the Jaccard similarity (token intersection over union) 
between $y_{w}$ and $y_{l}$ and the single-character Levenshtein edit distance between $y_{w}$ and $y_{l}$. The dataset with better minimal contrasts should result in a higher Jaccard similarity and a lower Levenshtein distance. Table~\ref{tab:data-contrastive} summarizes these analyses. By these measures, CLAIR delivers the best contrastive data by a wide margin.

\begin{table}[t!]
\centering
\begin{tabular}{@{} l c c @{}}
\toprule
Preference & Jaccard  & Levenshtein  \\ 
Dataset & ($\uparrow$ better) & ($\downarrow$ better) \\ \midrule
CLAIR           & \textbf{43.11}             & \textbf{1108}  \\
On-policy judge       & 39.06             &  1258 \\
Off-policy judge      & 18.05             &  1203 \\
\strongerpreferred       & 24.35             & 1607   \\ 
\bottomrule
\end{tabular}
\caption{Average token-level Jaccard similarity (intersection over union) 
and average character-level Levenshtein edit-distance between winning $y_w$ and losing $y_l$ answers for four comparable preference datasets built on top of \llama{}. The CLAIR dataset produces the best contrasts on both metrics.} 
\label{tab:data-contrastive}
\end{table}

\section{Anchored Preference Optimization} \label{sec:objectives}

\begin{figure*}[t]
    \centering    \includegraphics[width=\textwidth]{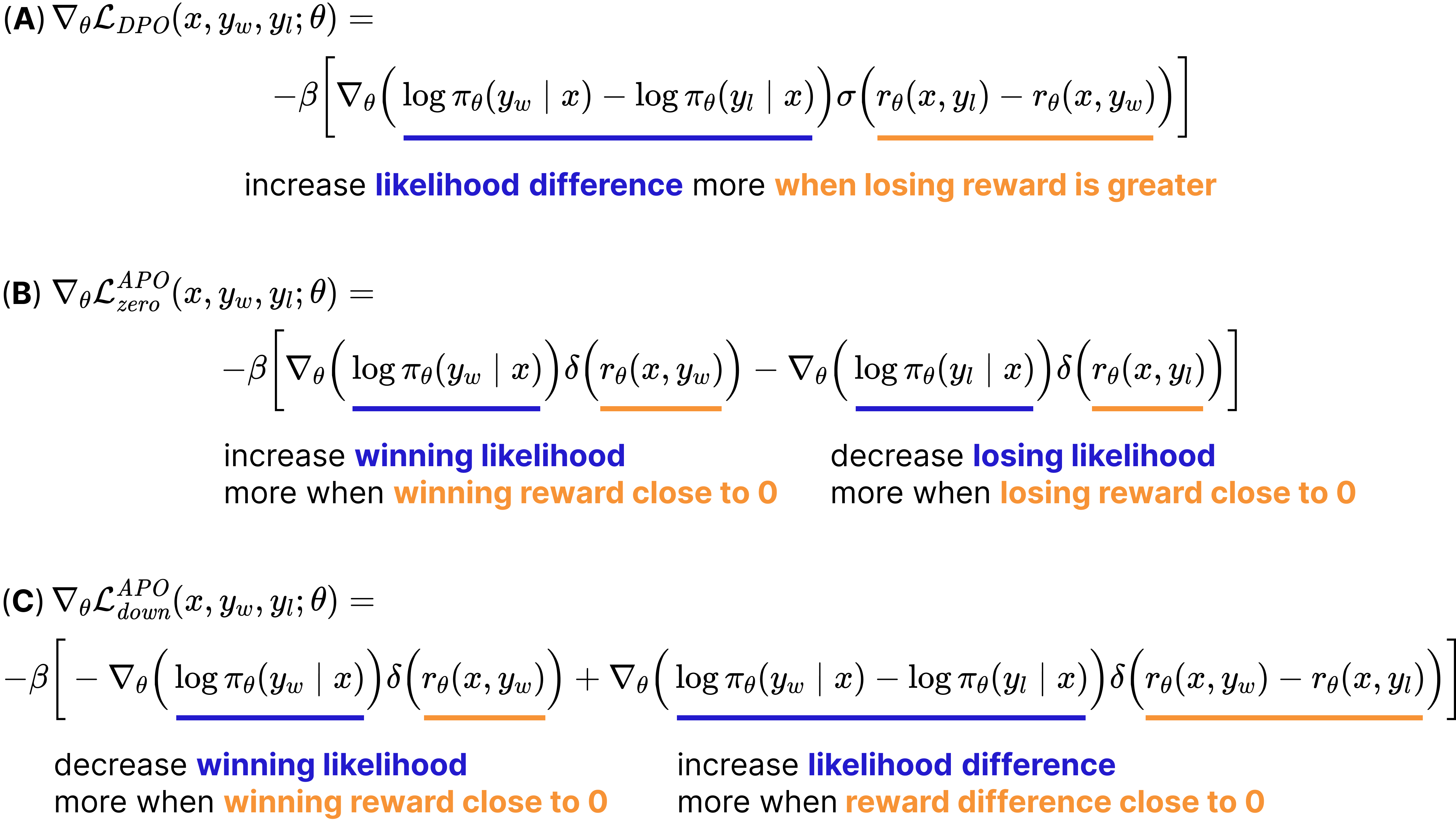}
    \caption{Comparison of gradients between DPO (equation~\textbf{A}), APO-zero (equation~\textbf{B}), and APO-down (equation~\textbf{C}). Each gradient term is decomposed in a \textcolor[HTML]{231ACD}{\textbf{direction}} and \textcolor[HTML]{F79336}{\textbf{magnitude}} factor. \textbf{Direction:} Either APO variant specifies explicitly if winning and losing likelihoods should increase or decrease during training. DPO only increases the likelihood difference, causing ambiguity with regard to the actual movement of these likelihoods during training. This explicit specification of direction is core to APO variants, and allows for a tighter fit between model and data during alignment. \textbf{Magnitude:} Each term in APO is scaled with a delta function. Here, $\delta(x) = \sigma(x)(1-\sigma(x))$ is a function with a global maximum at $x = 0$ that tends to $0$ for $x \to \pm \infty$. This causes APO gradients to saturate whenever the quantities being optimized have changed a lot compared to the beginning of training.  
    \citet{ethayarajh2024kto} theorize that such scaling leads to more robust optimization.
}
    \label{fig:apo-dpo-breakdown}
\end{figure*}

A preference triple $(x, y_{w}, y_{l})$ expresses the belief that $y_{w}$ is a more preferred output than $y_{l}$ for prompt~$x$. 
Alignment objectives use this relationship to align a model. Different objectives achieve this in very different ways, with deep consequences for the alignment process.

Direct Preference Optimization (DPO; \citealt{rafailov2024direct}) is a widely used and empirically successful alignment objective. The core stipulation of DPO is that the likelihood change of winning outputs during training needs to be greater than the likelihood change of losing outputs. This likelihood change for a prompt and output is denoted as the reward $r_\theta(x, y)$, which captures the log-ratio of likelihoods between the model during training $\pi_\theta(x\mid{}y)$ and the model before training, also called \textit{reference}, $\pi_{\text{ref}}(x\mid{}y)$:
\begin{equation}
r_\theta(x,y) = \beta \log \frac{\pi_{\theta}(y \mid x)}{\pi_{\text{ref}}(y \mid x)}%
\end{equation}
Here, $\beta$ is a hyperparameter which scales this log-ratio. This leads to the following DPO objective:
\begin{flalign}
\mathcal{L}_{\textit{DPO}}(x, y_w, y_l; \theta) &= \\ 
\nonumber -  \log \sigma &\Big( r_\theta(x,y_w) - r_\theta(x,y_l) \Big) 
\end{flalign}%

The DPO authors report that the gradient of this objective intuitively leads to an increased winning likelihood and decreased losing likelihood. However, this is only one possibility out of three distinct scenarios. Alternatively, DPO can increase the winning likelihood more than it increases the losing likelihood, or decrease the winning likelihood less than it decreases the losing likelihood \citep{feng2024towards}. 
These scenarios may end up producing vastly different models. 
As discussed in Section~\ref{sec:factors}, a winning output is not necessarily better than what the model produces \emph{before} alignment. 
In this case, DPO may hurt performance if it increases the likelihood of undesirable outputs.

To help researchers navigate these interactions, we introduce  Anchored Preference Optimization (APO). 
In essence, APO is a family of alignment objectives which offer fine-grained control over each of the rewards,
thus controlling the absolute increase or decrease in likelihood during training. 
In this paper, we focus in particular on variants that we call APO-zero and APO-down:
\begin{flalign}
\mathcal{L}_{\textit{zero}}^{\textit{APO}}(x, y_w, y_l; \theta) &= \\ 
\nonumber 
- \sigma \Big( r_\theta(x,&y_w)\Big) +  \sigma \Big( r_\theta(x,y_l) \Big) 
\end{flalign}
\begin{flalign}
\mathcal{L}_{\textit{down}}^{\textit{APO}}(x, y_w, y_l; \theta) &= \\ 
\sigma \Big(r_\theta(x,y_w)\Big) 
-  &\sigma \Big( r_\theta(x,y_w) - r_\theta(x,y_l) \Big)
\nonumber
\end{flalign}

APO-zero explicitly pushes for an increased likelihood of winning outputs and decreased likelihood of losing outputs during training. In contrast, APO-down decreases the likelihood of winning outputs and decreases the likelihood of losing outputs even more. 
If the model is better than the winning outputs ($y_w \prec \pi_\theta$), APO-down will intuitively be a better objective. If winning outputs are better than the model ($y_w \succ \pi_\theta$), APO-zero will be better.
Figure~\ref{fig:apo-dpo-breakdown} provides an interpretation of the gradients produced by both APO methods and compares these with DPO.

One can define additional APO objectives. In general, any contrastive objective 
(i.e., greater reward for winning outputs) 
which specifies additional constraints on either reward to achieve a tighter link between model and data (e.g., winning rewards should be positive) can be seen as a form of Anchored Preference Optimization. In Section~\ref{sec:related} we consider different alignment objectives and discuss how they relate to APO.  

One interesting variant of APO %
can be derived from the Kahneman--Tversky Optimization (KTO) objective of \citet{ethayarajh2024kto}. As originally defined, KTO does not operate on preference pairs, 
but rather requires only one unpaired answer and a label indicating if it was preferred or not; the goal of KTO is to push
the winning\,/\,losing reward above\,/\,below the Kullback–Leibler (KL) divergence between the model during training and the reference model. The APO perspective helps us see that there is a natural paired variant of KTO in which the KL-divergence functions as the anchor:
\begin{flalign}
\mathcal{L}_{\textit{KTO-pair}}(x, y_w, y_l; \theta) &= \\ \nonumber 
- \sigma \Big( r_\theta(x,y_w) - \beta~\textit{KL} & \Big) %
- \sigma \Big( \beta~\textit{KL} - r_\theta(x,y_l) \Big) \nonumber
\end{flalign}
This \textit{KL} term is non-negative, and 
thus the winning reward is pushed to be positive; the losing reward can still be either positive or negative. 

The KTO authors report that KTO leads to good alignment without an initial phase of Supervised Fine-Tuning (SFT) on the winning outputs, while DPO does benefit from this SFT phase in their experiments. APO sheds new light on this finding: an increase in likelihood of winning outputs is already built into KTO, whereas it is not guaranteed for DPO alone. However, this is only a desirable property of an alignment objective if the winning output quality is better than the target model's quality, as described in Section~\ref{sec:factors}. When aligning a strong model on preferences which contain generally lower quality outputs, a KTO-style objective runs the risk of deteriorating the model.

\section{Alignment Experiments}\label{sec:experiments}

To study the effectiveness of CLAIR and APO, we align \llama{} across the four comparable preference datasets described in Section~\ref{sec:data}, created from 32K \ultrafeedback{} prompts. We use \gptfour{} to act as \judge{} or \reviser{} when creating these datasets.
For every dataset, we align the model using the four different objectives described in Section~\ref{sec:objectives}. 
Additionally, we consider Supervised Fine-Tuning (SFT) on only the winning outputs as a baseline alignment objective.

\subsection{Evaluation Methodology}\label{sec:eval-method}

Human judgments are ultimately the best indicator of how well a model is aligned with human preferences. Chatbot Arena \citep{chiangchatbot} uses thousands of pairwise human judgements to produce a ranking of model performance. However, collecting these judgments can be prohibitively expensive. To overcome this obstacle, we measure model performance through benchmarks which correlate highly with this Chatbot Arena ranking.

\mixeval{} \citep{ni2024mixeval} is a benchmark with very high Chatbot Arena correlation  (0.96 rank correlation).
\mixeval{} features hard queries with known answers across a wide range of domains and uses a GPT3.5-turbo \citep{brown2020language, ouyang2022training} model to evaluate if predicted answers correspond with this ground-truth. This makes \mixeval{} more grounded in human knowledge and significantly cheaper to run compared to other popular evaluation frameworks such as \texttt{AlpacaEval} \citep{alpaca_eval, dubois2024length}. 
Under the hood, \mixeval{} utilizes queries sampled from 
\texttt{MATH} \citep{hendrycks2measuring}, 
\texttt{BBH} \citep{suzgun2023challenging}, 
\texttt{DROP} \citep{dua2019drop}, 
\texttt{GSM8k} \citep{cobbe2021training}, 
\texttt{AGIEval} \citep{zhong2024agieval}, 
\texttt{TriviaQA} \citep{joshi2017triviaqa}, 
\texttt{MBPP} \citep{austin2021program}, 
\texttt{MMLU}, \citep{hendrycksmeasuring}, 
\texttt{HellaSwag} \citep{zellers2019hellaswag}, 
\texttt{BoolQ} \citep{clark2019boolq}, 
\texttt{GPQA} \citep{rein2023gpqa}, 
\texttt{PIQA} \citep{bisk2020piqa}, 
\texttt{OpenBookQA} \citep{mihaylov2018can}, 
\texttt{ARC} \citep{clark2018think}, 
\texttt{CommonsenseQA} \citep{talmor2019commonsenseqa}, 
and \texttt{SIQA} \citep{sap2019social}.

Our evaluation of \llama{} before any additional alignment achieves a score of 41.45\% on the \texttt{2024-06-01} version of \mixeval{}. The gap between \llama{} and \gptfour{} is 17\%. On the \texttt{2024-08-11} split, \llama{} achieves 40.5\%.

Additionally, we consider the length-controlled \texttt{LC-AlpacaEval2.0} win rate \citep{dubois2024length}. However, two factors lead us to favor \mixeval{} as our primary evaluation tool. The first is practical: \texttt{LC-AlpacaEval2.0} is prohibitively expensive to run, we thus use \mixeval{} for the bulk of our evaluation. The second concerns the assessment itself: while both benchmarks are highly correlated with human-produced model rankings, \mixeval{} utilizes questions with known ground-truth answers whereas \texttt{LC-AlpacaEval2.0} uses an LLM judge without any ground-truth to decide correctness.

\begingroup
\renewcommand{\arraystretch}{.9}
\begin{table*}[tp]
\small
\centering
\begin{tabular}{@{} ll| rr|rr|rr @{}}
    \toprule
    \multicolumn{2}{c|}{} & \multicolumn{2}{c|}{\texttt{ME-Hard 2024-06-01}} & \multicolumn{2}{c|}{\texttt{ME-Hard 2024-08-11}} & \multicolumn{2}{c}{\texttt{LC-AlpacaEval2.0}} \\
    Dataset & Objective & Max $\Delta$ & Mean $\Delta$ & Max $\Delta$ & Mean $\Delta$ & Score $\Delta$ & Length $\Delta$ \\
    \midrule
    Judge & DPO & 1.10 & $-$0.74 { \scriptsize (1.15)} & 4.30 & 2.85 { \scriptsize (0.75)} & \underline{2.94} & $-$158  \\
    off-policy & KTO-pair & $-$1.00 & $-$2.89 { \scriptsize (0.96)} & 4.05 & 1.18 { \scriptsize (1.67)} & $-$5.69 & $-$437  \\
    
    & SFT & $-$1.95 & $-$1.63 { \scriptsize (1.06)} & 2.85 & 0.42 { \scriptsize (1.20)} & $-$22.29 & 12669 \\
    & APO-zero & 0.80 & $-$1.99 { \scriptsize (1.23)} & 4.65 & 1.26 { \scriptsize (1.62)} & $-$2.42 & $-$395 \\
    & \underline{APO-down} & \underline{2.70} & \underline{0.64} { \scriptsize (0.98)} & \underline{4.80} & \underline{3.52} { \scriptsize (0.85)} & 2.40 & $-$203 \\
    \midrule
    Judge & DPO & 4.00 & 0.56 { \scriptsize (1.61)} & 5.20 & 2.71 { \scriptsize (1.41)} & 4.98 & 341 \\
    on-policy & KTO-pair & 2.45 & $-$0.51 { \scriptsize (1.26)} & 5.05 & 1.13 { \scriptsize (1.70)} & 3.02 & 452  \\
    & SFT & 0.65 & $-$0.91 { \scriptsize (1.01)} & 4.20	 & 2.55	{ \scriptsize (0.70)} & 1.34 & 156 \\
    & \underline{APO-zero} & \underline{4.65} & 0.02 { \scriptsize (1.66)} & \underline{5.35} & 2.19 { \scriptsize (1.28)} & 5.51 & 484  \\
    & \underline{APO-down} & 3.65 & \underline{1.60} { \scriptsize (0.95)} & 4.25 & \underline{3.06} { \scriptsize (0.76)} & \textbf{7.63} & 386  \\
    \midrule
    \textbf{CLAIR} & DPO & 0.55 & $-$1.68 { \scriptsize (1.73)} & 5.05 & 2.77 { \scriptsize (1.40)} & 2.65 & 966  \\
    & KTO-pair & 2.15 & 0.79 { \scriptsize (0.98)} & 4.65 & 2.92 { \scriptsize (0.86)} & 4.33 & 160  \\
    & SFT & 0.65 & $-$0.91 { \scriptsize (1.01)}  & 2.70	& 0.92	{ \scriptsize (1.21)} & $-$0.47 & 6108 \\
    & \textbf{APO-zero} & \textbf{7.65} & \textbf{2.93} { \scriptsize (1.98)} & \textbf{5.95} & \textbf{4.39} { \scriptsize (0.89)} & \underline{5.08} & 520  \\
    & APO-down & $-$1.05 & $-$5.22 { \scriptsize (1.55)} & $-$1.20 & $-$3.61 { \scriptsize (1.05)} & $-$6.30 & 2559  \\
    \midrule
    Stronger & DPO & $-$5.00 & $-$6.94 { \scriptsize (1.03)} & $-$3.10 & $-$4.40 { \scriptsize (0.98)} & $-$2.89 & 597  \\
    Preferred & KTO-pair & $-$1.20 & $-$5.21 { \scriptsize (1.27)} & 2.25 & 0.50 { \scriptsize (1.13)} & 0.71 & 153  \\
    & \underline{SFT} & \underline{2.45} & \underline{0.49} { \scriptsize (1.31)} & \underline{5.05}	& \underline{2.73}	{ \scriptsize (1.21)} & \underline{6.99} & 1883 \\
    & APO-zero & $-$1.70 & $-$2.72 { \scriptsize (1.40)} & $-$4.85 & $-$12.02 { \scriptsize (5.38)} & 0.89 & 243  \\
    & APO-down & $-$6.50 & $-$12.51 { \scriptsize (4.97)} & 1.65 & 0.16 { \scriptsize (1.22)} & 1.87 & 10001 \\
    \bottomrule
\end{tabular}
\caption{Max and mean \mixeval{} improvements for the \texttt{2024-06-01} and \texttt{2024-08-11} splits, aggregated over 18 epochs of aligning \llama{}. Best overall performance \textbf{bold}, best performance per dataset \underline{underlined}, standard deviation in parentheses. 
While \mixeval{} functions as our primary evaluation tool, 
we also report the average \texttt{LC-AlpacaEval2.0} score increase over the two best \mixeval{} checkpoints, and average length increase (in characters) of the responses.
CLAIR leads to the greatest overall performance improvement on \mixeval{}. APO methods achieve the best performance across both Judged and CLAIR datasets. 
} \label{table:results}
\end{table*}
\endgroup

\subsection{Training Specifications} 
\llama{} is trained for a total of 18 epochs on each preference dataset and alignment objective, with a checkpoint saved every single epoch. 
The $\beta$ hyperparameter, common to all alignment objectives except SFT, is set to 0.1. 
Prompt and responses are truncated to 512 tokens each. 
Each model is trained using an effective batch size of 16 across one node of 8 \texttt{NVIDIA H100} GPUs, using the \texttt{RMSProp} optimizer with a learning rate of $2 \times 10^{-7}$, 
linearly decaying to 0 over the 18 epochs. 
All training is implemented using the \texttt{TRL} library \citep{vonwerra2022trl}.

\subsection{Results} \label{sec:results}

We report the maximal and mean \mixeval{} improvement over all checkpoints from the same training run. This helps us understand both the best-case and average impact of alignment across the entire training procedure.
We use both \texttt{2024-06-01} and \texttt{2024-08-11} versions of \mixeval{}, which each feature a distinct set of queries.
Due to the increased cost associated with \texttt{LC-AlpacaEval2.0}, we only measure the win rate for the two best \mixeval{} checkpoints and report their average. We use no system prompt for both evaluations.
Our analysis is summarized in Table~\ref{table:results} for every dataset and objective; we now discuss these results in more detail.

\subsubsection{Preference Data}
To assess the quality of a particular dataset, we consider the performance of that dataset when paired with its best objective. 
Using the APO-zero objective, \textbf{the contrastive CLAIR dataset leads to the greatest improvement}.
On the \texttt{2024-06-01} split of \mixeval{}, CLAIR leads to the greatest maximal improvement of +7.65\% and the greatest average improvement of +2.93\% out of all our experiments. This improvement of +7.65\% closes the relative gap with \gptfour{} by 45\% using only 32K pairs.

We noted in Section~\ref{sec:intro} that uncontrolled contrastiveness can degrade model performance. We see this dramatically in the results for the \strongerpreferred{} dataset, which can heavily degrade model performance. Like CLAIR, this dataset has all winning outputs produced by a stronger model. Unlike CLAIR, though, its examples provide no guarantee of relevant minimal contrasts. Thus, \textbf{the contrastiveness induced by the CLAIR revision process is a major driver of performance}.

Both on-policy judge and off-policy judge datasets lead to improved performance when paired with their best alignment objective, but \textbf{on-policy preferences lead to better performance compared to off-policy preferences}. This is intuitive; judgments about the target model's outputs are in general more relevant.

The \texttt{LC-AlpacaEval2.0} results generally follow a similar trend compared to \mixeval{}, although the on-policy judge dataset attains a higher score compared to CLAIR. While both benchmarks correlate highly with human ratings of models, \mixeval{} is our primary and most significant evaluation tool -- we are able to evaluate every model checkpoint across two \mixeval{} splits due to its low cost.
Additionally, we remark on a potential issue with the robustness of \texttt{LC-AlpacaEval2.0} in Appendix~\ref{app:lengthcontrol}.
A performance breakdown in function of \mixeval{}'s constituent benchmarks is given in Appendix~\ref{app:breakdown}.

\subsubsection{Alignment Objectives}
On \mixeval{}, \textbf{Anchored Preference Optimization (APO) consistently leads to the greatest performance increase for every preference dataset}, with the exception of the \strongerpreferred{} dataset, where all contrastive objectives underperform SFT. The relation between the preference dataset and the target model controls which variant of APO is best for any dataset, as predicted in Section~\ref{sec:factors}. \textbf{APO-down results in the best performance when winning outputs are generally worse than the target model}, as is the case for the off-policy judge dataset. 
\textbf{APO-zero is the best objective when winning outputs are generally better than the target model}, as is the case for CLAIR and on-policy judge datasets. The difference between alignment objectives is less salient for the on-policy judge dataset as compared to CLAIR, since winning on-policy judge outputs are only slightly better than \llama{} on average. Winning CLAIR outputs may be vastly better than \llama{} since they are produced by a stronger model, making the different in alignment objectives more noticeable.

\subsection{Analysis}

\begin{figure*}[t]
    \centering
    \includegraphics[width=\textwidth]{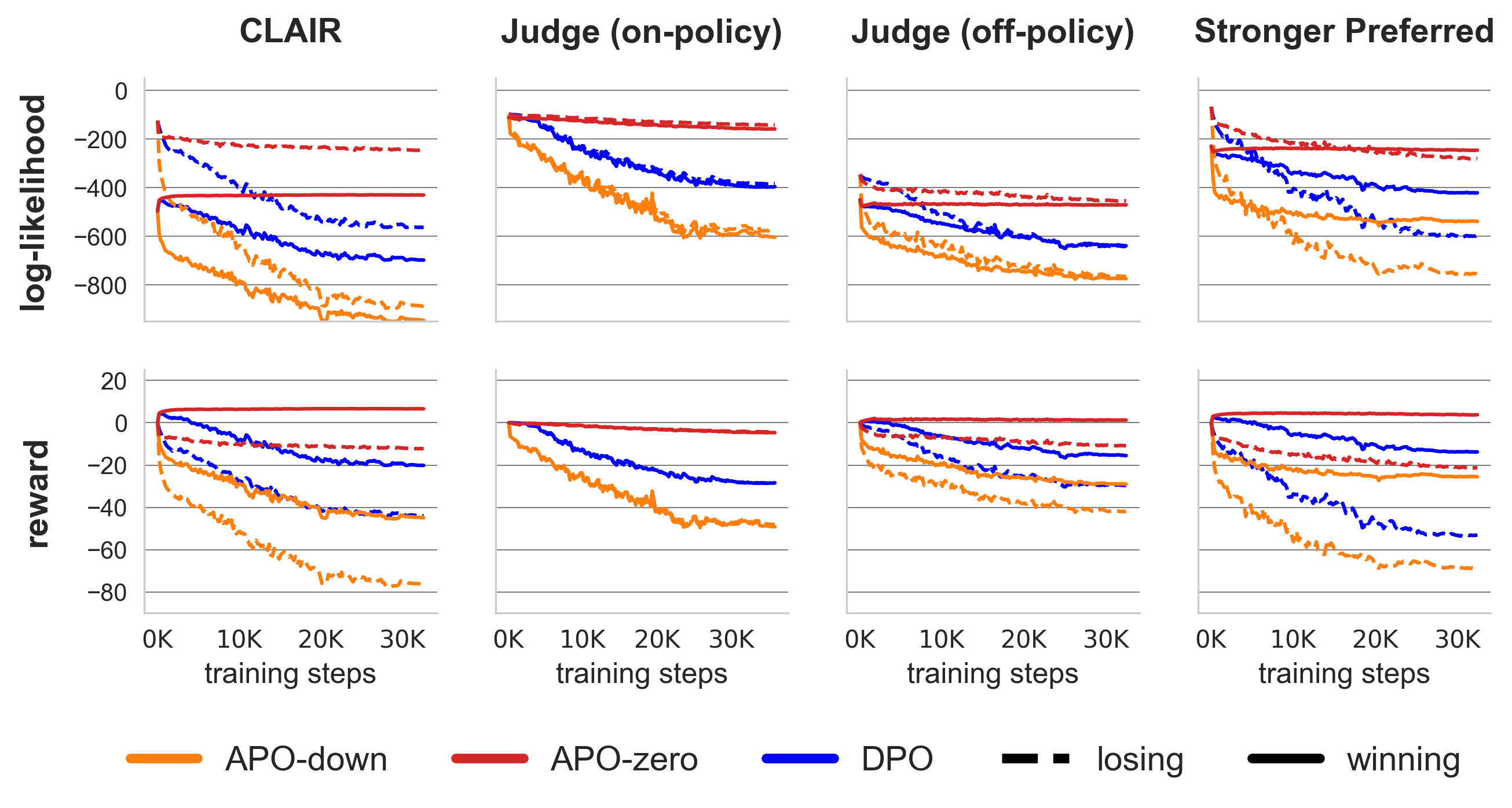}
    \caption{\textbf{Log-likelihood} and \textbf{reward} on held-out winning and losing outputs for \llama{} trained on \textbf{CLAIR}, 
    \textbf{on-policy judge}, \textbf{off-policy judge}, and \textbf{\strongerpreferred{}} preference datasets, using APO-down, APO-zero, or DPO alignment objectives.}
    \label{fig:training-curves}
\end{figure*}

To more deeply understand how the target model is changed during training, we can study the trajectories of winning\,/\,losing likelihoods and rewards on held-out preferences. Figure~\ref{fig:training-curves} plots these trajectories for the APO-down, APO-zero, and DPO experiments on each preference dataset, using 100 held-out preference pairs from that dataset.

\subsubsection{Preference Data}
First, we observe that the likelihoods help characterize the type of preference dataset. In the on-policy judge dataset, all answers are sampled from the target model and thus have a high likelihood. The off-policy variant has no answers coming from the target model, and hence all likelihoods are low. Both CLAIR and \strongerpreferred{} have losing outputs with high likelihood and winning outputs with low likelihood. 

Any initial discrepancy between log-likelihoods is normalized by the reward, which tracks changes in likelihood and thus starts at exactly 0.
The margin between winning and losing reward indicates how much more the winning likelihood increased during training. Positive reward margins can still produce negative log-likelihood margins, if any initial disparity between winning\,/\,losing log-likelihood is not overcome. This ends up being the case for our CLAIR dataset.

The training dynamics for CLAIR and \strongerpreferred{} look very similar, yet the downstream performance on \mixeval{} is completely different. This is because contrastive alignment objectives will exploit any difference between winning and losing outputs to decrease loss. Most of these differences in CLAIR are directly related to improving performance, because CLAIR itself is a minimally contrastive dataset. Many of the differences in \strongerpreferred{} may not be relevant.

\subsubsection{Alignment Objectives}
All three alignment objectives display systematic behavior across each dataset. APO-zero consistently leads to the greatest winning and losing rewards. APO-down consistently produces the lowest rewards. Both of these behaviors are as intended. DPO has a slightly more complicated dynamic, which is nonetheless consistent across datasets. In the initial steps of training, DPO tracks the behavior of APO-zero (high rewards) before following APO-down (low rewards) during the remainder of training. This explains why downstream DPO performance correlates most with APO-down. However, DPO is \emph{never} the best method on any dataset, because it falls between the distinct modes of APO-zero and APO-down.

Training models with contrastive alignment objectives is considerably more complex that conventional supervised fine-tuning. The result is dependent on the semantics of the alignment objective, the contrastive signal in the training data, and the relationship between data quality and target model. Our results show that paying attention to the interplay between these attributes is essential.

\section{Related Work} \label{sec:related}

We now characterize relevant alignment efforts and outline how they relate to Contrastive Learning from AI Revisions (CLAIR) and Anchored Preference Optimization (APO).

Reinforcement Learning from Human or AI Feedback (RLHF\,/\,RLAIF; \citealt{ouyang2022training, bai2022constitutional, yuanself}) is a technique used to align models with human preferences. Fundamentally, these approaches first train a reward model using preference judgments and subsequently optimize a Language Model for this reward using Reinforcement Learning \citep{schulman2017proximal}. To side-step the need for an explicit reward model, Direct Preference Optimization (DPO; \citealt{rafailov2024direct}) aligns an LM directly using a contrastive training objective.

We articulated two core insights concerning
\begin{enumerate*}[(i)]
    \item the role of contrastive preference data, and
    \item the need to anchor alignment depending on model and data.
\end{enumerate*}
These insights 
translate to any alignment effort which uses comparative preferences. 
For example, a reward model trained on spurious preference signals may be a less accurate proxy for real rewards, contributing to problems such as \emph{reward overoptimization} or \emph{hacking} \citep{gao2023scaling, rafailov2024scaling}.

For the remainder of this review, we first focus on contrastive alignment methods and their variants (of which \citealt{wang2024comprehensive} provide a detailed overview). Finally, we discuss related preference datasets and how they were created.

\paragraph{Changing the LM more\,/\,less:} \citet{amini2024direct} and \citet{wu2024beta} recognize that preference pairs can vary. Both works study \emph{how much more} preferred the winning output is, and seek to incorporate this into the objective by changing the model more\,/\,less depending on this preference strength. Using the difference in gold rewards as a substitute for preference strength, \citet{amini2024direct} add an instance-level margin to the contrastive objective while \citet{wu2024beta} scale the $\beta$ parameter at a batch-level. Other works also utilize a margin in the contrastive loss, but specify this as a static hyperparameter \citep{zhao2023slic, azar2024general,meng2024simpo}.
These contributions complement our own; they focus on \emph{how much} a model should change, whereas CLAIR creates better learning signals and APO more fully specifies the intended training dynamics.

\paragraph{Controlling training dynamics:} The tendency of DPO to decrease the winning likelihood has been remarked and analyzed in several works \citep{feng2024towards, pal2024smaug}. Some works use an additional loss term to explicitly increasing the likelihood of winning outputs \citep{hong2024reference, pentyala2024paft, adolphs2023cringe, zhao2023slic, xucontrastive}. While these methods can be seen as variants of Anchored Preference Optimization, they do not recognize the need to anchor the objective differently depending on dataset and model, and they do not offer methods that explicitly decrease the winning likelihood when required. Both \citet{rafailov2024scaling} and \citet{azar2024general} generalize a set of alignment methods, but neither allow for any anchoring.

\paragraph{Learning from unpaired data:} \citet{ethayarajh2024kto}, \citet{richemond2024offline}, and \citet{jung2024binary} use unpaired examples and rewards for alignment instead of paired examples. \citet{zhang2024negative} and \citet{duan2024negating} operate solely on undesirable examples in this unpaired setting. In contrast, our work exclusively operates on paired preferences.
However, the core insights of APO do apply to unpaired data. For example, \citet{ethayarajh2024kto} use binary desired\,/\,undesired labels for each answer. We argue this desirability is inherently relative to the model: the same example of desirable behavior used to improve a weak model may actually be an example of undesirable behavior compared to a stronger model, causing the need for anchoring.

\paragraph{Length-controlled optimization:} Preference pairs created through a judging paradigm can be biased towards preferring more verbose answers \cite{saito2023verbosity}. To prevent aligned models from inheriting this bias, \citet{meng2024simpo} and \citet{park2024disentangling} explicitly control for the length of generations during training. 
These constraints on generation length can be seamlessly integrated into APO methods as well. In addition, CLAIR revisions could further help with these efforts to reduce the verbosity bias. For example, the \reviser{} could be designed to not increase length.

\paragraph{Reference-free optimization:} Several objectives have opted to directly optimize the contrastive relation between winning\,/\,losing likelihoods instead of rewards, removing the need for a secondary reference model \citep{meng2024simpo, zhao2023slic, hong2024reference, xucontrastive}. Since all these methods are contrastive, the insights from CLAIR and APO directly apply. Additionally, the CLAIR dataset used in our experiments may shed light on the nature of reference-free optimization. Figure~\ref{fig:training-curves} shows that our models are sufficiently aligned on the CLAIR dataset when considering rewards, but the absolute likelihood of losing outputs is still greater. This is due to the initial discrepancy in likelihoods produced by the revision process.

\paragraph{Iterative optimization:}
Updating the reference model during training can improve results \citep{kim2024sdpo, rosset2024direct,wu2024self}. All of these insights are applicable to our work.

\paragraph{Preference Datasets:}

\citet{chiangchatbot} release a dataset of human preference judgements across conversations between humans and several AI assistants. To alleviate the need for human judges, some efforts focus on scaling preference annotations with LLM-based judges \citep{cui2024ultrafeedback, starling2023} or metric-based judges \citep{jiang2023llm}. Unlike our CLAIR method, these works do not create preferences through revisions. \citet{bai2022constitutional} use a set of predetermined criteria (called a \emph{constitution}) to prompt an LLM to revise answers and make them safer (see also \citealt{lambert2024self}). \citet{dubey2024llama} used human revisions in the development of the \texttt{llama-3.1} model family. While both efforts create preferences through revisions, we particularly focus on revisions that create a minimal contrast and studied the effect of this contrastiveness on alignment outcomes.

\section{Future work}

In this work, we have presented two variants of the APO objective family. Each method accounts for a distinct relationship between target model and preference pair during training. However, real world preference datasets may contain a wide range of different preference pairs, thus the dataset as a whole may not perfectly correspond with any single APO variant. To tackle this, a natural extension of APO could be to select the optimal APO variant at the preference pair level, instead of at the dataset level. Heuristically, this could be achieved using an off-the-shelf reward model to score each preference pair before training.

\section{Conclusion}
Alignment performance is significantly impacted by 
\begin{enumerate*}[(i)]
    \item the contrastiveness of the preference pairs and
    \item the relationship between target model and alignment data
\end{enumerate*}.
We introduce Contrastive Learning from AI Revisions (CLAIR), a data-creation method which produces better contrasting preference pairs, and Anchored Preference Optimization (APO), a family of alignment objectives with tailored training dynamics. Our experiments aligning \llama{} show that CLAIR preferences lead to the highest performance improvement out of four comparable preference datasets, and APO methods consistently outperform conventional alignment objectives.

\bibliography{custom,tacl2021}
\bibliographystyle{acl_natbib}

\newpage
\clearpage
\onecolumn

\appendix
\section{Preference Dataset Creation} \label{app:prompts}

\begin{table}
\centering
\begin{tabular}{p{0.1\linewidth} | p{0.9\linewidth}}
\toprule
Type & Prompt \\
\midrule
\reviser{} & You are a teacher and your task is to minimally improve a student's answer. I will give you a \{\{task\}\} and a \{\{student\_solution\}\}. Your job is to revise the \{\{student\_solution\}\} such that it is clearer, more correct, and more engaging. Copy all non-corrected parts of the student's answer. Do not allude to the \{\{corrected\_student\_solution\}\} being a revision or a correction in your final solution.\textbackslash{}n\textbackslash{}n\{\{task\}\}: 
\textbf{<instruction~$x$>}
\textbackslash{}n\textbackslash{}n\{\{student\_solution\}\}: 
\textbf{<losing output~$y_l$>}
\textbackslash{}n\textbackslash{}n-----------------\textbackslash{}n\textbackslash{}nLet's first think step by step with a \{\{teacher\_reasoning\}\} to decide how to improve the \{\{student\_solution\}\}, then give the \{\{corrected\_student\_solution\}\}. Mention the \{\{teacher\_reasoning\}\} and \{\{corrected\_student\_solution\}\} identifiers to structure your answer.\textbackslash{}n\textbackslash{}n \\
\judge{}
& You are a teacher and your task is to pick the best student's answer. The best answer is the most clear, most correct, and most engaging answer. I will give you a \{\{task\}\} and \{\{student\_solution\_1\}\} and \{\{student\_solution\_2\}\}. Your final answer must contain [1] if \{\{student\_solution\_1\}\} was best, else [2].\textbackslash{}n\textbackslash{}n\{\{task\}\}: 
\textbf{<instruction~$x$>}
\textbackslash{}n\textbackslash{}n\{\{student\_solution\_1\}\}: 
\textbf{<first output~$y_1$>}
\textbackslash{}n\textbackslash{}n\{\{student\_solution\_2\}\}: 
\textbf{<second output~$y_2$>}
\textbackslash{}n\textbackslash{}n-----------------\textbackslash{}n\textbackslash{}nLet's first think step by step with a \{\{teacher\_reasoning\}\} to decide which solution is better, and then answer [1] or [2].\textbackslash{}n\textbackslash{}n \\
\bottomrule
\end{tabular}
\caption{
Prompt templates used for creating preference triples $(x, y_l, y_w)$ with the \reviser{} and \judge{} 
function of Equation~\ref{eq:clair} and \ref{eq:rlaif}. The variables in the prompt template are \textbf{bolded} and bracketed. Both prompts target clear, correct, and engaging outputs. The \reviser{} prompt instructs that a losing output $y_l$ should be minimally improved to create the winning output $y_w$. Instead, the \judge{} prompt picks the winning\,/\,losing output out of two candidates $y_1$ \& $y_2$. Both prompts also instruct a model to produce a reasoning before revising or judging.
} \label{tab:prompts}
\end{table}

\subsection{Prompts}
The prompts we use for the \reviser{} and \judge{} function of Equation~\ref{eq:clair} and \ref{eq:rlaif} are given in Table~\ref{tab:prompts}. Both prompts contain instructions to prefer more clear, more correct, and more engaging outputs. The \reviser{} prompt creates a preference pair by minimally revising and improving an output according to these preferences. Instead, the \judge{} prompt selects a more preferred output given two candidate answers.

\subsection{Preference Pair Filtering}
We reject revisions or judgments if the LLM failed to follow formatting guidelines specified in the revising or judging prompt. Additionally, we reject revisions if they altered the length of the original output too much; we found this mainly happens when the LLM misunderstands the revision prompt. Starting from the same 32K instructions sampled from \ultrafeedback{}, this procedure creates 29K CLAIR pairs, 29K \strongerpreferred{} pairs, 29K off-policy Judge pairs, and 32k on-policy Judge pairs.
We adapted the code by \citet{Williams2023} to efficiently query closed-source LLMs in parallel over API.

\clearpage
\section{\mixeval{} Performance Breakdown} \label{app:breakdown}
\mixeval{} features queries from a wide range of established benchmarks, as outlined in Section~\ref{sec:eval-method}. Previously, we reported on the overall \mixeval{} performance. Table~\ref{tab:breakdown} breaks down this overall performance in function of these different benchmarks. While \mixeval{} often incorporates only a few queries from any given benchmark, the overall performance correlates highly with human judgements.

\begin{table}[tp]
\centering
\begin{tabular}{lr|r|rrrr}
\toprule
  \mixeval{} split & \# query
  & \texttt{Llama-3-8B} & + CLAIR & + Judge  & + Judge  & + Stronger  \\
 \multicolumn{2}{r|}{} & \texttt{-Instruct} & & (on-policy) &  (off-policy) &  Preferred \\
\midrule
Overall score & 988 & 41.45 & \textbf{49.10} & 46.10 & 44.15 & 43.90 \\
\midrule
TriviaQA & 267 & 34.30 & \textbf{49.20} & 42.40 & 43.70 & 39.80 \\
MMLU & 231 & \textbf{43.70} & 39.00 & 42.00 & 36.80 & 34.60 \\
DROP & 167 & 50.20 & 58.70 & 64.30 & \textbf{64.90} & 58.90 \\
AGIEval & 71 & 31.00 & 38.00 & 38.00 & \textbf{39.40} & 38.00 \\
HellaSwag & 61 & 29.50 & \textbf{37.70} & 26.20 & 29.50 & 27.90 \\
CommonsenseQA & 50 & 60.00 & \textbf{72.00} & 60.00 & 48.00 & 58.00 \\
BoolQ & 37 & 40.50 & \textbf{45.90} & 32.40 & 21.60 & 27.00 \\
GSM8k & 22 & 60.00 & 80.00 & 69.50 & 63.20 & \textbf{84.10} \\
SIQA & 20 & 45.00 & \textbf{50.00} & 40.00 & 15.00 & 40.00 \\
MATH & 16 & 47.50 & 63.70 & 51.30 & 58.80 & \textbf{73.10} \\
BBH & 16 & 51.30 & \textbf{68.80} & 57.50 & 60.60 & 66.90 \\
OpenBookQA & 8 & 62.50 & 62.50 & 50.00 & 62.50 & \textbf{75.00} \\
GPQA & 8 & 12.50 & 25.00 & 25.00 & 25.00 & \textbf{37.50} \\
PIQA & 8 & 50.00 & 62.50 & 62.50 & 62.50 & \textbf{75.00} \\
ARC & 4 & \textbf{0.00} & \textbf{0.00} & \textbf{0.00} & \textbf{0.00} & \textbf{0.00} \\
MBPP & 2 & \textbf{0.00} & \textbf{0.00} & \textbf{0.00} & \textbf{0.00} & \textbf{0.00} \\
\midrule
\multicolumn{2}{l|}{Objective used:} & / & APO-zero & APO-zero &  APO-down &  SFT \\
\bottomrule
\end{tabular}
\caption{Breakdown of \mixeval{} performance (version \texttt{2024-06-01}) in function of which dataset the queries originate from. Analysis given for \llama{} and our best models on the CLAIR, Judge (on-policy), Judge (off-policy), and \strongerpreferred{} datasets. While individual splits may not always indicate the best model (particularly when the amount of queries is low), the overall score correlates highly with human judgments about model performance (Chatbot Arena Elo; \citealt{chiangchatbot}). \mixeval{} uses a GPT3.5-turbo model to rate if a response to a query agrees with a known gold-truth response.} \label{tab:breakdown}

\end{table}

\clearpage
\section{Unpaired APO}
In this work, we designed datasets and alignment objectives for paired preferences (output $y_l \prec y_w$ for input $x$). The original KTO objective \citep{ethayarajh2024kto} was designed to operate on desirability data (output $y$ for input $x$ was desirable or not), which does not use such paired preferences. We consider an unpaired variant of our APO-zero loss, called APO-zero-unpaired, which resembles the KTO objective but which fixes the KL term to zero.
Table~\ref{table:results-unpaired} compares KTO with APO-zero-unpaired, keeping everything else comparable with our main results in Table~\ref{table:results}. To turn our paired datasets into unpaired datasets, we turn each datapoint consisting of two outputs into two datapoints with one output.

There is no clear winner between KTO and APO-zero-unpaired across the board. Within each dataset however, there always is a clear winner. This reflects the main findings of our work, different alignment objectives have distinct semantics, and different datasets require different semantics. APO-zero-unpaired consistently trains faster, due to not calculating the KL term. In some cases, the KTO objective can take 60\% longer to train.

\begingroup
\renewcommand{\arraystretch}{1.0}
\begin{table*}[tp]
\small
\centering
\begin{tabular}{@{} ll| rr|rr |r @{}}
    \toprule
    \multicolumn{2}{c|}{} & \multicolumn{2}{c|}{\texttt{ME-Hard 2024-06-01}} & \multicolumn{2}{c|}{\texttt{ME-Hard 2024-08-11}} &  \\
    Dataset & Objective & Max $\Delta$ & Mean $\Delta$ & Max $\Delta$ & Mean $\Delta$ & Train Time \\
    \midrule
    Judge & KTO & \underline{2.10}  &  \underline{$-$2.70} { \scriptsize (1.67)} & \underline{4.75} & \underline{1.31} { \scriptsize (1.61)} & 19h 18m 10s \\
    off-policy & APO-zero-unpaired & $-$0.40  & $-$3.67 { \scriptsize (1.68)} & 4.35 &  0.66 { \scriptsize (1.44)} & 12h 32m 58s \\
    
    \midrule
    Judge & KTO & 3.50  &  1.28 { \scriptsize (1.11)} & 4.85 & 2.70 { \scriptsize (1.35)} & 19h 40m 10s \\
    on-policy & APO-zero-unpaired & \textbf{4.35} & \underline{1.31}  { \scriptsize (1.44)} & \underline{5.60}  & \underline{3.92} { \scriptsize (0.99)} & 13h 49m 55s  \\
    \midrule
    CLAIR & KTO & \underline{3.75}  &  \textbf{1.47} { \scriptsize (1.39)} & \textbf{5.80} &  \textbf{4.12} { \scriptsize (1.09)}  & 17h 33m 24s \\
    & APO-zero-unpaired & 1.40 & $-$1.49 { \scriptsize (1.77)} & 3.20 & 1.13 { \scriptsize (1.21)} & 12h 31m 03s  \\
    \midrule
    Stronger & KTO & $-$3.25 &  $-$4.73{ \scriptsize (1.01)} & 0.30 &  $-$1.18{ \scriptsize (0.75)} & 19h 07m 29s \\
    Preferred & APO-zero-unpaired & \underline{$-$2.70}  & \underline{$-$4.57}  { \scriptsize (1.32)} & \underline{2.95} &  \underline{0.50} { \scriptsize (1.25)} & 12h 38m 49s  \\
    \bottomrule
\end{tabular}
\caption{Max and mean \mixeval{} improvements for the \texttt{2024-06-01} and \texttt{2024-08-11} splits, aggregated over 18 epochs of aligning \llama{}. Best overall performance \textbf{bold}, best performance per dataset \underline{underlined}, standard deviation in parentheses. 
KTO is the best unpaired loss given the off-policy Judge and CLAIR datasets, while APO-zero-unpaired performs better when given the on-policy Judge and Stronger Preferred datasets.
KTO can take 60\% longer to train for the same configuration.
} \label{table:results-unpaired}
\end{table*}
\endgroup

\clearpage
\section{How well does \texttt{AlpacaEval} control for response lengths?} \label{app:lengthcontrol}
GPT4 as a judge is known to favor more verbose responses, which can artificially inflate \texttt{AlpacaEval} win rates for verbose models \citep{dubois2024length}. To counteract this bias, \citet{dubois2024length} estimate a length-controlled \texttt{AlpacaEval} win rate, which we report on in Table~\ref{table:results}. Specifically, the authors adopt a causal inference framework to answer the question "What would the \texttt{AlpacaEval} metric be, if the outputs of all models had the same length as those of the baseline?" \citep{dubois2024length}.

In order to meaningfully apply causal inference, a few key assumptions need to be met. The \emph{Positivity} assumption \citep{hernan2006estimating} states that, when estimating the effect of a treatment, there are at least some subjects which receive the treatment for all covariates.
Intuitively, the Positivity assumption applied to the length-control question states that you need to observe at least some long and some short responses for every model in order to accurately estimate how the response length influences the model's win rate.

The \texttt{AlpacaEval} framework does not check if this Positivity assumption is met, potentially giving bad estimates for the length-controlled win rates in some settings. If a certain model consistently generates responses longer than those of the baseline, it is impossible to accurately estimate how good the responses would be if they were as long as the baseline.

This may give us insights into some of our length-controlled \texttt{AlpacaEval} win rates. For example, the SFT result on the Stronger Preferred dataset in Table~\ref{table:results} seems disproportionately high in comparison to the \mixeval{} results for that same experiment. This model is considerably more verbose than \llama{}, as evident from the large response length increase associated with this experiment ( + 1883 characters on average). It is possible the Positivity constraint was not met for this experiment, causing the length-controlled framework of \texttt{AlpacaEval} to provide inaccurate estimates.

While a more thorough study of length-controlled win rate is out of scope for this work, one potential avenue towards a more robust length-controlled win rate would be to specifically prompt models to generate shorter or longer answers if the Positivity constraint is not met.

\end{document}